\documentclass[journal]{IEEEtran}
\usepackage[utf8]{inputenc}
\usepackage{comment}
\usepackage{graphicx}
\usepackage{booktabs}
\usepackage{stfloats}
\usepackage{multirow}
\usepackage[table]{xcolor}
\usepackage{tabularx}
\definecolor{headerblue}{RGB}{226,238,248}
\definecolor{bestgreen}{RGB}{232,245,233}
\usepackage{subcaption}
\usepackage{cite}
\usepackage{amsmath,amssymb,amsfonts}
\usepackage{algorithmic}
\usepackage{graphicx}
\usepackage[table]{xcolor}
\usepackage{textcomp}
\usepackage{booktabs}
\usepackage{float}
\usepackage{hyperref}
\hypersetup{
    colorlinks=true,
    linkcolor=blue,
    citecolor=blue,
    urlcolor=blue
}

\begin{document}

\title{\textit{PRIME}: Evaluating Prompt Resolution Under Incompatible Instructions in LLMs}
%\author{\IEEEauthorblockN{Anonymous Authors}}
% Add Author names here
\author{
\IEEEauthorblockN{
Tehreem Javed$^{*}$,
Shumaim Fatimah$^{*}$,
Masooma Bakhtiari$^{*}$,
Gibrail Islam,
Mehwish Fatima
}

\IEEEauthorblockA{
School of Electrical Engineering and Computer Science (SEECS),\\
National University of Sciences and Technology (NUST),\\
Islamabad, Pakistan\\
\{tjaved.msds24seecs,
sfatimah.msds24seecs,
mbakhtiari.msds24seecs, \\
gibrail.islam,
mehwish.fatima\}@seecs.edu.pk
}

\IEEEauthorblockA{
$^{*}$Equal contribution \\
$^{\dagger}$Corresponding author: mehwish.fatima@seecs.edu.pk
}

}
\maketitle

\begin{abstract}
Large language models (LLMs) often encounter conflicting prompts, although current instruction following benchmarks assess those meta-instructions in isolation, limiting the insights about how models process conflicting instructions.  We introduce a framework \textit{PRIME}(\textit{Prompt Resolution under Incompatible Meta-Instructions Evaluation}) to analyze behavior of LLMs when provided with conflicting instructions. \textit{PRIME} purposefully produces calibrated conflicts across response length, output format, and reasoning; classifying model responses with a deterministic behavioral taxonomy. We are evaluating five instruction tuned open weight LLMs  in two distinct settings, balanced and naturally distributed. The conclusion we reach upon analysis is that conflict type is more significant in affecting behavior than model scale, and various failure modes across different categories of conflict. Our findings emphasize the value of developing conflict awareness and suggest ability of LLM to follow instructions cannot be assessed through isolated constraints alone. 
\end{abstract}


\section{Introduction}

Large language models (LLMs) function as chatbots, coding assistants,  search engines, and decision-making tools. Users often send prompts with conflicting or irrelevant instructions.  For example: User asks LLM to synthesize solution to a problem step by step and demands to output the final answer only. A single response cannot meet both requirements at the same time. As shown in Table~\ref{tab:reasoning-example}, despite the instructions, either not to show intermediate steps or just to show the final answer, Mistral outputs final answer while Qwen and TinyLlama show intermediate steps. Typically, models do not identify such contrasts. They mainly follow one command, none, or give irrelevant outcomes. These errors reduce reliability in the outputs of various applications such as clinical decision support systems, legal documents, and automatic grading.

Current research on instruction following mainly investigates consistent prompts. Instruction tuning, supervised fine-tuning, and reinforcement learning from human feedback enhance compliance with user requests, while benchmarks such as IFEval~\cite{Zhou2023} and FollowBench~\cite{Jiang2024} assess adherence to individual or compatible constraints. More recently, ConInstruct~\cite{He2026} introduced conflicting instructions and demonstrated that models seldom recognize contradictions or request clarification. Current assessments provide little insight into how models behave  when conflicts arise. It remains unclear whether models systematically favor one directive over another, whether they ignore both instructions, and whether these behaviors vary across conflict types and model families. As a result, current instruction-following benchmarks provide only a partial view of model behavior in realistic prompting scenarios. 

\begin{table}[t]
\caption{An example of reasoning conflict resolution across models.}
\label{tab:reasoning-example}
%\centering
\scriptsize
\begin{tabularx}{\columnwidth}{X}
\toprule
\rowcolor{blue!12}
\textbf{Base instruction:}
If $x + 5 = 17$, what is $x$? Think through the problem step by step.
Return only the final result without explanation.\\ \midrule
\textbf{Constructed prompt:} Solve step by step. Only final answer. \\
\textbf{Question:} If $x + 5 = 17$, what is $x$? Think through the problem step by step.
Return only the final result without explanation.\\
\midrule
\rowcolor{red!10}
\textbf{Mistral:} Answer: $x = 12$\\
\midrule
\rowcolor{blue!12}
\textbf{Qwen:} $x = 12$.
To solve for $x$, subtract 5 from both sides: $x + 5 - 5 = 17 - 5$,
which simplifies to $x = 12$. The correct answer is $\mathbf{12}$.\\
\midrule
%\rowcolor{red!10}
\textbf{TinyLlama:} Step 1: subtract 5 from both sides.
$x + 5 - 5 = 17 - 5$, so $x = 12$. The answer is $\mathbf{12}$.\\
\bottomrule
\end{tabularx}
\end{table}

To address this gap, we propose \textit{PRIME} (Prompt Resolution under Incompatible Meta-Instructions Evaluation), a framework for analyzing how LLMs respond to contradictory meta-instructions. \textit{PRIME} introduces controlled conflicts involving reasoning depth, response length, and output format, and categorizes responses using a deterministic behavioral taxonomy. Rather than measuring instruction compliance alone, the framework characterizes conflict-resolution behavior through directive preference and instruction neglect. We evaluate five instruction-tuned open-weight LLMs under both balanced and naturally distributed settings. Our in-depth analysis shows that conflict type plays a stronger role than model scale in shaping response behavior and reveals consistent resolution patterns across models. The contributions of this work are as follows:

\begin{itemize}
\item We propose \textit{PRIME} for evaluating prompt resolution under incompatible meta-instructions.
\item We introduce a deterministic behavioral taxonomy for characterizing responses to contradictory meta-instructions.
\item We provide an empirical analysis of conflict-resolution behavior across reasoning, length, and formatting conflicts.
\item We show that conflict type influences model behavior more strongly than model scale and exposes distinct failure modes across model families.
\end{itemize}

%=======================================================
\section{Related Work}
\label{sec:related}
We divide this section into three categories: instruction tuning, instruction-following evaluation, and contradictory instruction analysis. %These categories establish the foundations of instruction following, but leave open the question of how models behave when instructions conflict.

\subsection{Instruction Tuning}
Instruction tuning aims to align language models with natural language instructions. FLAN~\cite{Wei2022} demonstrates that training on diverse instructions improves zero-shot generalization, while InstructGPT~\cite{Ouyang2022} shows that reinforcement learning from human feedback improves instruction-following behavior. Subsequent studies report further gains through larger instruction datasets and model scaling~\cite{Chung2022,Zhang2023survey}. These works focus on learning to follow instructions, assuming that user meta-instructions are internally consistent.

\subsection{Instruction-Following Evaluation}
Several benchmarks evaluate instruction adherence. IFEval~\cite{Zhou2023} measures compliance with verifiable constraints, while FollowBench~\cite{Jiang2024} extends evaluation to fine-grained and multi-level instructions. These benchmarks assess whether a model follows specified requirements, but they do not examine situations where satisfying one instruction requires violating another.

\subsection{Contradictory Instructions}
A few recent studies explore instruction conflicts. \cite{Zhao2021,Liu2024,Zheng2023} investigate how prompt ordering for instruction position can influence model decisions. ConInstruct~\cite{He2026} introduces contradictory instructions and reports that models rarely identify or explicitly acknowledge conflicts. However, existing work provides limited insight into how models resolve competing meta-instructions. It remains unclear whether models consistently favor one instruction, ignore both, or exhibit different behaviors across conflict types.

\begin{figure}[ht]
    \centering
    \includegraphics[width=0.8\columnwidth]{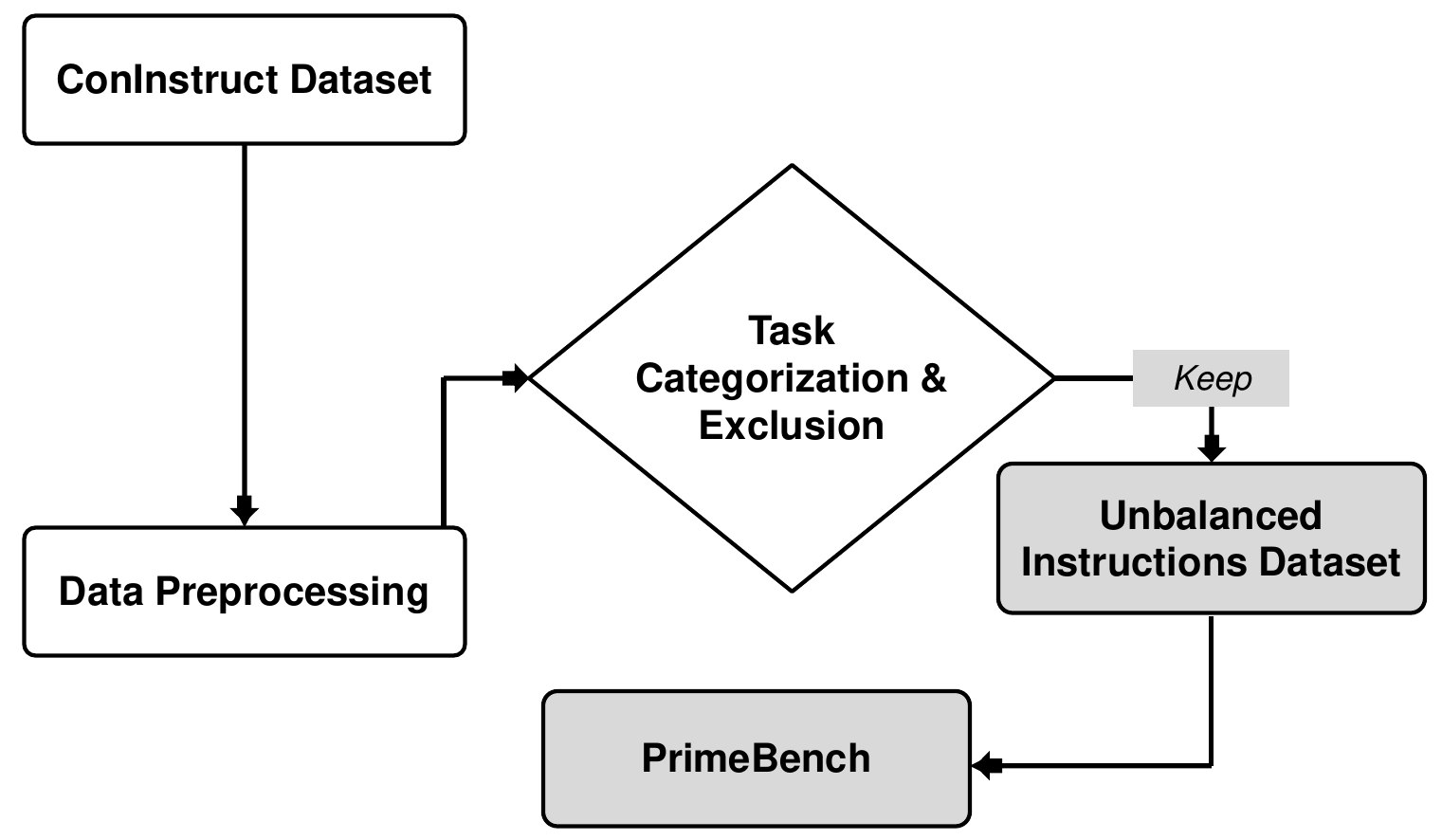} 
    \caption{Overview of \textit{PRIMEBench} construction process.}
    \label{fig:1}
\end{figure}
%==========================================
\section{Dataset Construction}
We introduce \textit{PRIMEBench} to evaluate how LLMs resolve contradictory instructions. Unlike existing instruction-following benchmarks that focus on compatible constraints, \textit{PRIMEBench} contains prompts where satisfying one meta-instruction necessarily violates another. \textit{PRIMEBench} supports systematic analysis of conflict-resolution behavior across different instruction types. Figure~\ref{fig:1} summarizes the dataset construction process.

%============================
\subsection{Source Instructions}
We construct \textit{PRIMEBench} from ConInstruct~\cite{He2026}, a publicly available instruction-following dataset containing diverse natural language tasks. We remove duplicate, empty, and excessively long instructions and exclude highly open-ended tasks such as storytelling, email writing, and travel planning. These tasks make it difficult to evaluate compliance against well-defined instruction constraints.

The remaining instructions are grouped into three categories: (i) \textbf{Arithmetic}: instructions involving numerical computation, (ii) \textbf{Logic}: instructions requiring conditional or deductive reasoning, and (iii) \textbf{Conceptual}: instructions requiring explanations or knowledge-based responses. This categorization enables us to examine whether conflict-resolution behavior varies across different reasoning requirements.

\begin{figure}[t]
    \centering
    \includegraphics[width=\columnwidth]{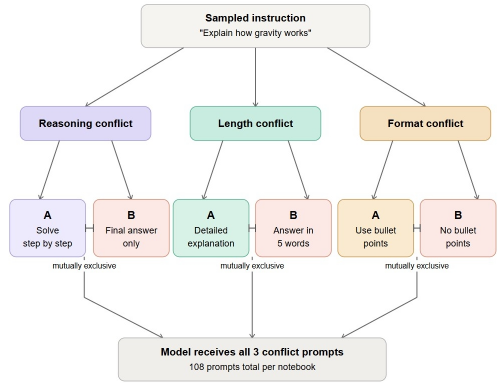}
    \caption{Instruction Conflict Types}
    \label{fig:2}
\end{figure}
%============================
\subsection{Conflict Generation}
For each source instruction, we generate three prompt variants by injecting incompatible meta-instructions. We consider three common conflict types: (i) \textbf{Reasoning Conflict}: step-by-step explanation versus final-answer-only response, (ii) \textbf{Length Conflict}: detailed explanation versus strict brevity, and (iii) \textbf{Format Conflict}: bullet-point versus non-bullet-point output. Each conflict introduces a situation in which both meta-instructions cannot be satisfied simultaneously. Figure~\ref{fig:2} illustrates the conflict-generation process.

\begin{table}[ht]
\centering
\caption{Composition of \textit{PRIMEBench} under balanced and unbalanced evaluation settings.}
\label{tab:2}
\begin{tabular}{lcccc}
\toprule
\textbf{Setting} & \textbf{Arithmetic} & \textbf{Logic} & \textbf{Conceptual} & \textbf{Total} \\
\midrule
Unbalanced & 18 & 34 & 20 & 72 \\
Balanced   & 12 & 12 & 12 & 36 \\
\bottomrule
\end{tabular}
\end{table}

\subsection{Benchmark Composition}
\textit{PRIMEBench} contains 72 source instructions distributed across arithmetic, logic, and conceptual reasoning tasks. Applying three conflict types to each instruction yields 216 contradictory prompts. To assess the effect of category distribution, we construct two evaluation settings: (i) an unbalanced set that preserves the natural distribution of task categories and (ii) a balanced subset containing an equal number of instructions from each category. Table~\ref{tab:2} summarizes the composition of \textit{PRIMEBench}.

\begin{figure}[ht]
    \centering
    \includegraphics[width=\columnwidth]{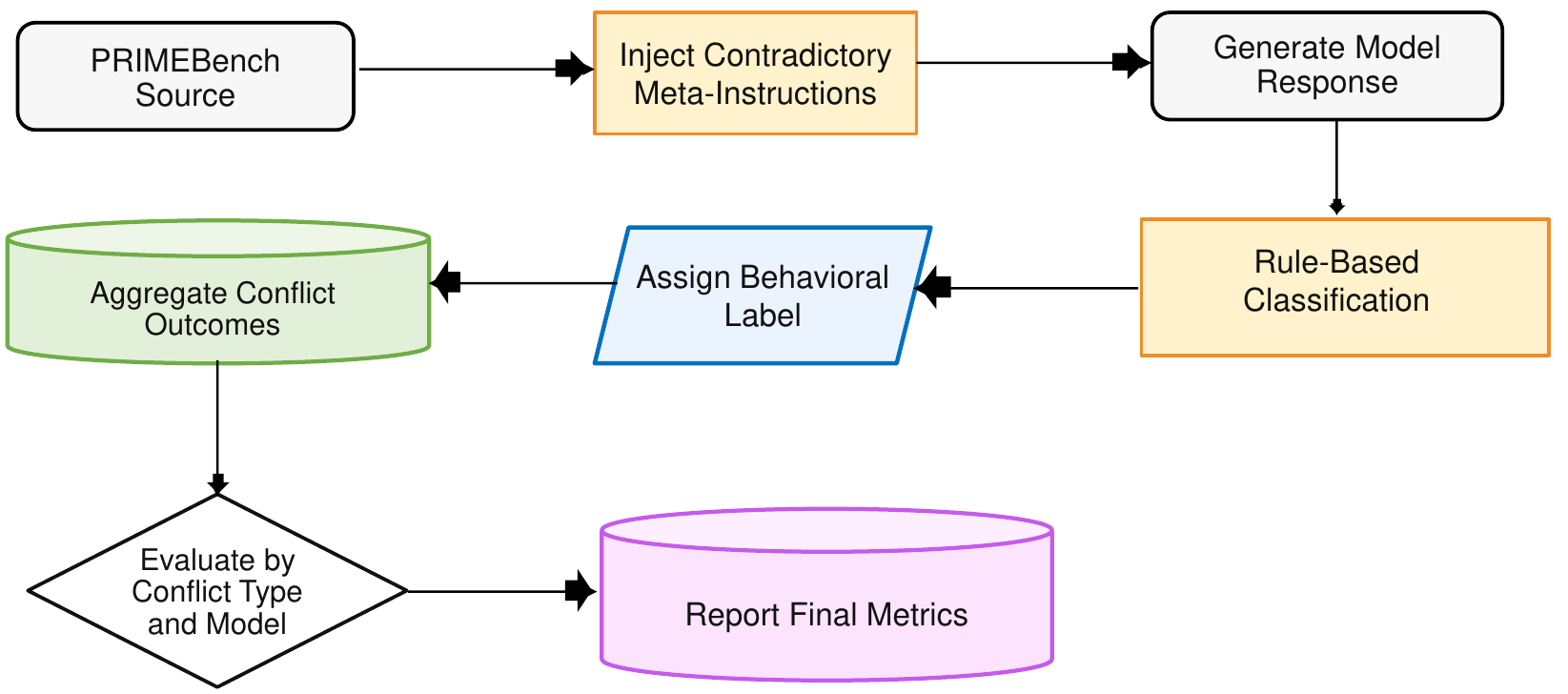}
    \caption{Overview of the \textit{PRIME} framework.}
    \label{fig:prime_framework}
\end{figure}

%=====================================================================
\section{\textit{PRIME} Framework}\label{sec:framework}
We evaluate how LLMs respond when instructions become mutually incompatible with PRIME. Given a source instruction from \textit{PRIMEBench}, the framework injects a pair of contradictory meta-instructions, collects the model response, and assigns a behavioral label through deterministic rules. Figure~\ref{fig:prime_framework} presents the overall workflow.
We investigate how LLMs resolve conflicting instructions within a prompt. For this, we generate three conflict types: (i) reasoning, (ii) length, and (iii) format. Figure~\ref{fig:2} illustrates the process of injecting conflict types.

\subsection{Behavioral Classification}
\textit{PRIME} classifies each response according to its adherence to the conflicting meta-instructions. We define four behavioral labels: (i) \textit{Follows A}, where the response satisfies only the first directive, (ii) \textit{Follows B}, where the response satisfies only the second directive, (iii) \textit{BOTH}, where the response satisfies both meta-instructions, and (iv) \textit{NEITHER}, where the response satisfies neither directive.

The classification rules depend on the conflict type. For reasoning conflicts, \textit{Follows A} requires a step-by-step explanation, whereas \textit{Follows B} requires only the final answer. For length conflicts, \textit{Follows A} requires a detailed response, whereas \textit{Follows B} requires strict brevity. For format conflicts, \textit{Follows A} requires a bullet-point structure, whereas \textit{Follows B} requires its absence.

\subsection{Behavioral Outcomes and Evaluation Metrics}
To analyze conflict-resolution behavior, we aggregate the response labels into three higher-level outcomes: (i) \textit{First Bias}, corresponding to \textit{Follows A}, (ii) \textit{Second Bias}, corresponding to \textit{Follows B}, and (iii) \textit{Ignoring}, corresponding to \textit{NEITHER}. These outcomes capture whether a model systematically favors one directive or fails to follow either instruction.

We report two aggregate metrics. Instruction Adherence Rate (IAR) measures the proportion of responses that satisfy at least one directive, while None Rate (NR) measures the proportion of responses that satisfy neither directive. Together, these metrics provide a concise summary of model behavior under contradictory instructions.

%========================================================================
\section{Experimental Design}\label{sec:method}
We investigate instruction-following behavior under conflicting meta-instructions as follows. 
\subsection{Models}
We evaluate five instruction-tuned open-weight LLMs: TinyLlama-1.1B~\cite{Zhang2024}, Qwen2-1.5B~\cite{Yang2024}, Gemma-2B~\cite{Gemma2024}, StableLM-3B~\cite{StableLM}, and Mistral-7B~\cite{Jiang2023}. The models span a range of parameter sizes and architectures~\cite{Wang2025SLM}, allowing us to examine whether conflict-resolution behavior varies across model families and scales.

\subsection{Inference Configuration}
We evaluate all models on \textit{PRIMEBench} using greedy decoding. We generate a maximum of 200 new tokens for each prompt and remove the prompt tokens from the generated output before classification. We apply the same inference configuration to all models to ensure comparability.

\subsection{Behavioral Analysis}
We analyze model behavior from three perspectives. First, we examine the relationship between conflict type and behavioral outcome. Second, we compare conflict-resolution patterns across models. Third, we investigate whether response length is associated with directive preference and instruction neglect. Responses are grouped into four length categories: Very Short (0--50 characters), Short (51--150), Medium (151--500), and Long (501--1000).

\subsection{Statistical Analysis}
We use Pearson's chi-square test to evaluate associations between experimental factors and behavioral outcomes. Statistical significance is assessed at $\alpha = 0.05$. To quantify effect size, we report Cram{e}r's $V$ for all significant associations.

\begin{table}[ht]
\centering
\caption{Overall conflict-resolution outcomes across evaluation settings.}
\label{tab:overall_results}
\begin{tabular}{lcc}
\toprule
\textbf{Metric} & \textbf{Balanced} & \textbf{Unbalanced} \\
\midrule
IAR & 70.22 & 71.33 \\
NR  & 29.78 & 28.67 \\
\midrule
First Bias   & 48.44 & 50.00 \\
Second Bias  & 21.78 & 21.33 \\
Ignoring     & 29.78 & 28.67 \\
\bottomrule
\end{tabular}
\end{table}

%====================================================================
\section{Results}\label{sec:results}
\subsection{Overall Conflict-Resolution Behavior}
The general behavioral outcomes in balanced and unbalanced evaluation settings are summarized in Table~\ref{tab:overall_results}. Results are quite uniform: In most instances, models respond to at least one of the meta-instructions and about one third of the responses are to none of the instructions. First Bias emerges as the dominant outcome in both settings, indicating a general tendency to favor one meta-instruction rather than explicitly identify or resolve the contradiction. The similarity between the two settings suggests that the observed behaviors are not driven by category distribution only.

\subsection{Impact of Conflict Type}
\begin{figure*}[hb]
\centering
\begin{subfigure}[t]{0.48\textwidth}
    \centering
    \includegraphics[width=\textwidth]{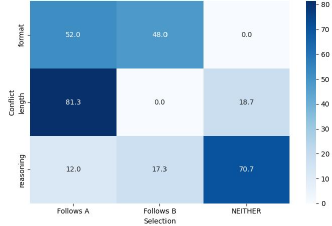}
    \caption{Balanced evaluation setting}
    \label{fig:conflict_bal}
\end{subfigure}
\hfill
\begin{subfigure}[t]{0.48\textwidth}
    \centering
    \includegraphics[width=\textwidth]{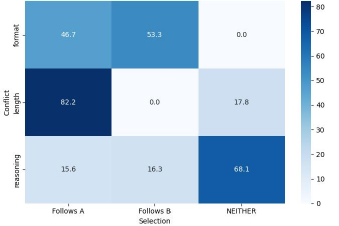}
    \caption{Unbalanced evaluation setting}
    \label{fig:conflict_unbal}
\end{subfigure}
\caption{Behavioral outcomes across conflict types. Reasoning conflicts consistently produce the highest levels of instruction neglect, whereas format conflicts are typically resolved by selecting one of the competing directives. Length conflicts exhibit a strong preference for detailed responses over strict brevity constraints.}
\label{fig:conflict_comparison}
\end{figure*}

The most significant difference in model behavior is by conflict type. Table~\ref{tab:conflict_results} and Figure~\ref{fig:conflict_comparison} illustrate that format conflicts are typically resolved by choosing one of the conflicting meta-instructions, which leads to high adherence and a small amount of instruction neglect. In length conflicts, there is a strong preference for detailed answers, which indicates that at times models may not be constrained by length and are inclined to provide a lengthy response. Reasoning conflicts, on the other hand, have the highest percentage of \textit{NEITHER} responses. This pattern shows that models have difficulties in inferring and not inferring intermediate reasoning. These trends are consistent in both evaluation settings, underscoring the key role of conflict type in affecting model behavior.

\begin{table}[t]
\centering
\caption{Behavioral outcomes across conflict types.}
\label{tab:conflict_results}
\begin{tabular}{lccc}
\toprule
\textbf{Conflict} &
\textbf{Follows A} &
\textbf{Follows B} &
\textbf{Neither} \\
\midrule
\multicolumn{4}{c}{\textbf{Balanced}} \\ \midrule
Format    & 52.00 & 48.00 & 0.00 \\
Length    & 81.33 & 0.00  & 18.67 \\
Reasoning & 12.00 & 17.33 & 70.67 \\
\midrule
\multicolumn{4}{c}{\textbf{Unbalanced}}\\ \midrule
Format    & 53.00 & 47.00 & 0.00 \\
Length    & 82.00 & 0.00  & 18.00 \\
Reasoning & 15.00 & 17.00 & 68.00 \\
\bottomrule
\end{tabular}
\end{table}

%---------------------------------
\subsection{Model-Specific Resolution Patterns}
\begin{figure*}[htb]
\centering

\begin{subfigure}[t]{0.48\textwidth}
    \centering
    \includegraphics[width=\textwidth]{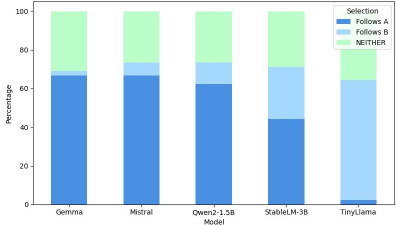}
    \caption{Balanced evaluation setting}
    \label{fig:model_bal}
\end{subfigure}
\hfill
\begin{subfigure}[t]{0.48\textwidth}
    \centering
    \includegraphics[width=\textwidth]{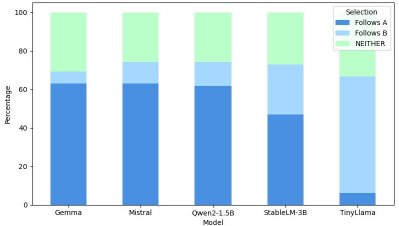}
    \caption{Unbalanced evaluation setting}
    \label{fig:model_unbal}
\end{subfigure}
\caption{Conflict-resolution behavior across models. Gemma, Mistral, and Qwen2-1.5B predominantly favor the first directive, whereas TinyLlama consistently favors the second. These preferences remain stable across evaluation settings.}
\label{fig:model_comparison}
\end{figure*}

Table~\ref{tab:model_results} and Figure~\ref{fig:model_comparison} demonstrate behavioral distributions for individual models. Gemma and Mistral almost always choose the first prompt, while TinyLlama strongly leans towards the second prompt. StableLM-3B shows a comparatively even distribution for behavioral outcomes. Significantly, the rating of these preferences is largely unchanged between the two types of evaluation (balanced and unbalanced), as presented in Figure~\ref{fig:model_comparison}, although the task composition differs between the two types of evaluation. The stability implies that the directive preference is a property of the model, not a distribution property of the data sets. Significantly, models can have similar behavior across different parameter scales, and models of similar size can be different in adopting conflict resolutions. The observations in this study suggest that conflict resolution behavior is related to model specific alignment tendencies, not only to the size of the parameters. The findings also indicate that instruction tuned models can have significantly different strategies for the same set of conflicting meta-instructions.

\begin{table}[t]
\centering
\caption{Conflict-resolution behavior across models.}
\label{tab:model_results}
\begin{tabular}{lccc}
\toprule
\textbf{Model} &
\textbf{First Bias} &
\textbf{Second Bias} &
\textbf{Ignoring} \\
\midrule

\multicolumn{4}{c}{\textbf{Balanced}} \\
\midrule
Gemma       & 66.67 & 2.22  & 31.11 \\
Mistral     & 66.67 & 6.67  & 26.67 \\
Qwen2-1.5B  & 62.22 & 11.11 & 26.67 \\
StableLM-3B & 44.44 & 26.67 & 28.89 \\
TinyLlama   & 2.22  & 62.22 & 35.56 \\

\midrule
\multicolumn{4}{c}{\textbf{Unbalanced}} \\
\midrule
Gemma       & 65.00 & 3.33  & 31.67 \\
Mistral     & 65.00 & 6.67  & 28.33 \\
Qwen2-1.5B  & 65.00 & 10.00 & 25.00 \\
StableLM-3B & 50.00 & 25.00 & 25.00 \\
TinyLlama   & 5.00  & 61.67 & 33.33 \\

\bottomrule
\end{tabular}
\end{table}

\begin{table}[ht]
\centering
\caption{Association between experimental factors and behavioral outcomes.}
\label{tab:stats}
\begin{tabular}{lccc}
\toprule
\textbf{Relationship} &
\textbf{$\chi^2$} &
\textbf{$p$-value} &
\textbf{Cramer's $V$} \\
\midrule

\multicolumn{4}{c}{\textbf{Balanced}} \\
\midrule
Conflict $\times$ Selection & 145.75 & $1.66\times10^{-30}$ & 0.569 \\
Model $\times$ Selection    & 78.03  & $1.22\times10^{-13}$ & 0.416 \\

\midrule
\multicolumn{4}{c}{\textbf{Unbalanced}} \\
\midrule
Conflict $\times$ Selection & 184.86 & $6.75\times10^{-39}$ & 0.555 \\
Model $\times$ Selection    & 98.52  & $8.58\times10^{-18}$ & 0.405 \\

\bottomrule
\end{tabular}
\end{table}

\subsection{Statistical Significance}
Table~\ref{tab:stats} presents the results of the statistical analysis. In both evaluation settings, there are significant relationships between conflict type and behavioral outcome, and between model identity and behavioral outcome. The impact of the conflict type is consistently greater than that of the impact of conflict identity. In quantitative terms, suggesting that the nature of the conflict has a significant effect on behavior than the choice of identity for that conflict. These findings validate qualitative trends across the analysis.

\subsection{Effect of Evaluation Setting}
The outcome distributions for the balanced and unbalanced evaluations are very similar. There are small differences between settings in terms of instruction adherence, directive preference, and instruction neglect. The stability of the behavioral patterns found in \textit{PRIME} supports the notion that these patterns are stable across varying category distributions and are mostly shaped by the nature of the conflicting meta-instructions themselves.

%===================================================
\section{Discussion}\label{sec:discussion}
We assess conflict resolution behavior using \textit{PRIME}: by adding controlled contradictions in reasoning, length, formatting of instructions. The type of conflict serves as the strongest predictor of behavior among all experiments. Reasoning conflicts produce the  highest rate of instruction neglect, while format conflicts tend to resolve through model selecting one of the conflicting directives. When length constraints apply, models consistently prefer to write detailed answers rather than producing very short ones. These differences show that the nature of the contradiction, not instruction following itself, drives conflict resolution.
Requests that expose or conceal intermediate reasoning differ qualitatively from formatting or length requests and yield significantly higher rates of instruction neglect. We observe consistent directive preferences across models: Gemma, Mistral, and Qwen2‑1.5B follow the first directive, while TinyLlama follows the second. StableLM exhibits a more balanced distribution of outcomes. These preferences remain stable across the balanced and unbalanced evaluation settings. Models with different parameter scales often exhibit similar behavioral patterns, while models of comparable size can adopt opposing conflict-resolution strategies. This observation suggests that conflict-resolution behavior is influenced more by model-specific alignment characteristics than by parameter scale alone. Taken together, the results indicate that conflict type primarily determines the difficulty of a contradiction, whereas model identity determines the strategy used to resolve it.

The balanced and unbalanced evaluations produce highly similar distributions of adherence, directive preference, and instruction neglect. This consistency indicates that the observed patterns are not artifacts of category imbalance. Instead, they arise from the interaction between contradictory meta-instructions and model-specific response tendencies. The stability of these patterns across evaluation settings suggests that the behavioral trends identified by \textit{PRIME} are robust to changes in task distribution.

Current instruction-following benchmarks primarily evaluate whether models satisfy individual constraints. \textit{PRIME} evaluates a different capability: how models behave when no response can satisfy all constraints simultaneously. The results show that instruction compliance and conflict resolution are related but distinct aspects of model behavior. Evaluating both provides a more complete assessment of instruction-following systems operating in realistic prompting environments.

\subsection{Key Takeaways}
(1) Conflict type influences behavioral outcomes more strongly than model scale, (2) reasoning conflicts produce substantially higher instruction neglect than length or format conflicts, (3) directive preferences remain stable across evaluation settings and reflect model-specific alignment tendencies, and (4) instruction following and conflict resolution represent distinct capabilities that require separate evaluation.

%=============================
\section{Conclusions}
\label{sec:conclusion}
We present \textit{PRIME}, a framework for evaluating how large language models respond to contradictory instructions. Using \textit{PRIMEBench}, we introduce controlled conflicts in reasoning, length, and formatting constraints and analyze model behavior through a deterministic conflict-resolution taxonomy.

Our results show that conflict type influences behavior more strongly than model scale. Reasoning conflicts generate the highest rates of instruction neglect, while length and format conflicts exhibit distinct directive preferences. These findings suggest that instruction following and conflict resolution are related but distinct capabilities. \textit{PRIME} provides a simple framework for evaluating this overlooked aspect of model behavior.  
\bibliographystyle{IEEEtran}
\bibliography{references}
\end{document}